\documentclass[12pt]{article}
\usepackage{algpseudocode}
\usepackage{amsmath, amssymb, amsthm}
\usepackage{makecell}
\usepackage{float, fullpage, graphicx, multirow,parskip, subcaption, setspace}
\usepackage{comment}
\usepackage{url}
\usepackage{enumitem}
\usepackage{tikz}
\usepackage{bm, bbm}
\usetikzlibrary{shapes.geometric, positioning}
\usetikzlibrary{quotes, angles}
\usepackage{rotating}
\usepackage{hyperref}
\usepackage{natbib}
\usepackage{placeins}
\usepackage{algorithm2e}
\RestyleAlgo{ruled}

\definecolor{SkyBlue}{RGB}{14, 118, 188}
\definecolor{BrightRed}{RGB}{223,82, 78}

\usepackage{amsmath,amsfonts,bm}

\def\eqref#1{equation~\ref{#1}}
\def\1{\bm{1}}

\DeclareMathAlphabet{\mathsfit}{\encodingdefault}{\sfdefault}{m}{sl}
\SetMathAlphabet{\mathsfit}{bold}{\encodingdefault}{\sfdefault}{bx}{n}

\newcommand{\E}{\mathbb{E}}

\newcommand{\R}{\mathbb{R}}

\hypersetup{pdfborder = {0 0 0.5 [3 3]}, colorlinks = true, linkcolor = BrightRed, citecolor = SkyBlue}

\def\keywordname{{\bfseries \emph Keywords}}%
\def\keywords#1{\par\addvspace\medskipamount{\rightskip=0pt plus1cm
\def\and{\ifhmode\unskip\nobreak\fi\ $\cdot$
}\noindent\keywordname\enspace\ignorespaces#1\par}}

\newcommand{\codeloc}{https://github.com/YunyiShen/Perceiver-diffusion-autoencoder}

\usepackage{cleveref}
\usepackage{placeins, float}

\title{Diffusion Autoencoders with Perceivers for Long, Irregular and Multimodal Astronomical Sequences}
\author{Yunyi Shen \thanks{EECS, MIT \texttt{yshen99@mit.edu}} \and Alexander Gagliano \thanks{IAIFI, MIT; CfA, Harvard, \texttt{gaglian2@mit.edu}}}

\begin{document}
\def\bY{\bm{Y}}
\def\by{\bm{y}} % vector of all observations

\def\bz{\bm{z}}
\def\bX{\bm{X}}
\def\bx{\bm{x}} % vector of single set of covariates

\def\R{\mathbb{R}}
\def\N{\mathcal{N}}
\def\P{\mathbb{P}}
\def\E{\mathbb{E}}

\def\Xcal{\mathcal{X}}

\maketitle

\begin{abstract}
Self-supervised learning has become a central strategy for representation learning, but the majority of architectures used for encoding data have only been validated on regularly-sampled inputs such as images, audios. and videos. In many scientific domains, data instead arrive as long, irregular, and multimodal sequences. To extract semantic information from these data, we introduce the Diffusion Autoencoder with Perceivers (daep). daep tokenizes heterogeneous measurements, compresses them with a Perceiver encoder, and reconstructs them with a Perceiver-IO diffusion decoder, enabling scalable learning in diverse data settings. To benchmark the daep architecture, we adapt the masked autoencoder to a Perceiver encoder/decoder design, and establish a strong baseline (maep) in the same architectural family as daep. Across diverse spectroscopic and photometric astronomical datasets, daep achieves lower reconstruction errors, produces more discriminative latent spaces, and better preserves fine-scale structure than both VAE and maep baselines. These results establish daep as an effective framework for scientific domains where data arrives as irregular, heterogeneous sequences.
\end{abstract}

\section{Introduction}
\label{sec:intro}
Self-supervised learning (SSL) has emerged as a powerful paradigm for representation learning, driving major advances in language, vision, and audio domains \citep{jing2020self}. In the canonical framing of SSL techniques, however, data are defined on regular grids such as pixels in images or fixed-rate samples in audio and video.

In many scientific and real-world applications, data instead arrive as long sequences sampled irregularly in time. Biomedical records contain patient measurements obtained at uneven intervals \citep{krishnan2022self}, sensor networks report readings based on event triggers or battery constraints, and in astrophysics, photometry is obtained with uneven temporal sampling due to observational constraints. Developing SSL methods that can natively handle these irregular, multimodal inputs is an important and open challenge across domains.

Astronomical surveys provide a particularly valuable context for investigating SSL. Spectroscopic surveys like the Sloan Digital Sky Survey \citep{york2000sloan} and the Dark Energy Spectroscopic Instrument Survey \citep{abareshi2022overview} produce spectra for millions of sources, while time-domain imaging surveys like the Zwicky Transient Facility Bright Transient Survey \citep{bellm2018zwicky}, the Asteroid Terrestrial-impact Last Alert System \citep{2018Tonry_ATLAS}, and the Young Supernova Experiment \citep{jones2021young} produce similar volumes of irregularly-sampled photometry. SSL methods that combine these data modalities are now beginning to emerge \citep{zhang2024maven, rizhko2025astrom3}.

Reconstruction-based methods have become a popular SSL paradigm, following their success in natural language processing and vision \citep{gui2024survey, chen2021empirical, rani2023self}. Diffusion models achieve state-of-the-art sampling in the image domain \citep{ho2020denoising, dhariwal2021diffusion}, but operate directly in pixel space without learning compressed representations \citep{kwon2022diffusion}. Diffusion autoencoders \citep{preechakul2022diffusion} address this by coupling a deterministic encoder with a diffusion decoder, simultaneously enabling both representation learning and high-fidelity generation. Diffusion autoencoders have achieved impressive performance in the image domain, but an architecture does not yet exist to extend the capabilities to to long, irregular, and multimodal sequences.

In this paper, we introduce the Diffusion AutoEncoder with Perceiver (daep), an architecture designed for self-supervised learning of long, irregular, and multimodal sequences. Our proposed architecture combines three components: (i) a Perceiver encoder that flexibly accommodates variable-length and tokenized inputs across modalities; (ii) a compact latent bottleneck for representation learning; and (iii) a Perceiver-IO diffusion decoder that reliably reconstructs the original sequence regardless of the regularity of its sampling. This design allows daep to scale to datasets with millions of heterogeneous samples, while producing both high-fidelity reconstructions and semantically-structured latent spaces.

To contextualize our approach, we also develop a masked autoencoder baseline using the same Perceiver backbone (maep), enabling a controlled comparison between masking-based and diffusion-based objectives in the irregular-sequence regime. Across unimodal and multimodal spectroscopic and photometric astronomical datasets, daep achieves comparable or lower reconstruction error compared to VAE and maep baselines, with particular improvements in reconstructing critical high-frequency data features. While motivated by astronomical data, the proposed architecture is domain-agnostic and can be used for representation learning in healthcare, finance, and other areas where irregular and multimodal sequences are obtained.

\section{Background}
\label{sec:background}
\textbf{Diffusion models.}
Diffusion models are score-based generative models that achieve state-of-the-art performance in image and video generation. These models learn a denoising process that transforms random noise into samples drawn from a target data distribution. Generation proceeds by iteratively denoising samples drawn from a Gaussian prior $x_T\sim\mathcal{N}(0, I)$ into a clean data sample $x_0$ after $T$ denoising steps. \citet{ho2020denoising} proposed to learn a noise model $\epsilon_{\theta}(x_t, t)$ that predicts the noise added at diffusion time $t$ to the corrupted data $x_t$. The model is trained by minimizing $||\epsilon_{\theta}(x_t, t)-\epsilon_t||_2^2$, where $\epsilon_t$ is the true noise added to clean data $x_0$ to produce $x_t$. During generation, the corruption process is inverted to produce a trajectory $x_T, x_{T-1}, \dots, x_{0}$, typically with large $T$. \citet{song2020denoising} introduced a deterministic variant, the denoising diffusion implicit model (DDIM), which enables generation in fewer steps using the trained noise prediction model. When conditioning variables $z$ are available, the noise model can be extended as $\epsilon_{\theta}(x_t, z, t)$ and trained with the same loss $||\epsilon_{\theta}(x_t, z, t)-\epsilon_t||_2^2$.

\textbf{Diffusion autoencoders.}
Diffusion autoencoders were originally proposed for the image domain by \citet{preechakul2022diffusion}. They encode data and reconstruct it using a conditional diffusion model. Because the encoding guides every denoising step, diffusion autoencoders capture fine-grained detail more effectively than, for example, variational autoencoders \citep{kingma2013auto}. However, the original design relied on U-Nets \citep{preechakul2022diffusion, dhariwal2021diffusion}, which are better suited to regular modalities such as images. Formally, the model encodes tokenized data into a latent representation $z=\text{Enc}_{\theta}(x)$, and a conditional score model $\epsilon_{\theta}(x_t, z, t)$ decodes data via a diffusion process. Training minimizes the score-matching loss $||\epsilon_{\theta}(x_t, \text{Enc}_{\theta}(x), t)-\epsilon_t||_2^2$.

\textbf{Perceiver.}
Perceiver and Perceiver-IO \citep{jaegle2021perceiver, jaegle2021perceiverio} provide a general framework to (1) encode irregularly sampled sequences into a latent representation and (2) query outputs from this latent space. This makes them a natural fit for integration with diffusion transformers \citep{dhariwal2021diffusion}. Perceiver-IO is designed to accommodate diverse data settings for both encoding and decoding, and scales linearly with the input data. We provide additional details on the perceiver architecture in Section~\ref{sec:method} below.

\textbf{Masked autoencoders.}
Masked autoencoders (MAEs) \citep{he2022masked} are an alternative approach to representation learning. Instead of reconstructing the full data from the latent alone, MAEs mask a subset of the input and predict the masked values conditioned on both the latent representation and the unmasked portion. Formally, the data are split into a masked portion $x_{\text{m}}$ and an unmasked portion $x_{\text{u}}$. The latent representation is obtained by encoding the unmasked portion, $z=\text{Enc}_{\theta}(x_{\text{u}})$, while the decoder learns a function $\text{Dec}_{\theta}(x_{\text{u}}, z)$ by minimizing the masked reconstruction loss $||x_{\text{m}}-\text{Dec}_{\theta}(x_{\text{u}}, \text{Enc}_{\theta}(x_{\text{u}}))||_2^2$. This model is not a fully generative decoder, since it requires access to the unmasked data tokens for conditioning rather than decoding from positional information alone. 

\textbf{Related work.}
Autoencoding and dimensionality reduction have a long history in representation learning. Early models that remain widely used include variational autoencoders \citep[VAEs;][]{kingma2013auto} and their variants, such as hierarchical VAEs \citep{vahdat2020nvae} and Vector-Quantized VAEs \citep{van2017neural, razavi2019generating}. These models are still common in physics applications, though they often suffer from posterior collapse \citep{van2017neural, higgins2017beta} and are generally less expressive than GANs \citep{goodfellow2020generative} or diffusion models \citep{ho2020denoising}.

Researchers have explored combining VAEs with diffusion models to improve generative quality, for example by learning a diffusion prior \citep{wehenkel2021diffusion} or training diffusion models on VAE latent spaces \citep{kwon2022diffusion, yan2021videogpt}. Beyond VAEs, masked autoencoders \citep[MAEs;][]{he2022masked} have recently gained attention as efficient learners for images and videos, but primarily for regularly sampled modalities. MAEs reconstruct masked regions from unmasked context, a strategy well suited to modalities with strong local structure such as images or audio (diffusion models can also be adapted to perform masked reconstruction; \citealt{wei2023diffusion}); this strategy is less effective for data with long-range dependencies, particularly when sequences are sparse. Despite their impressive performance in image and audio domains, these methods struggle to encode high-frequency structure in irregularly-sampled sequences with a large number of tokens. 

\section{Diffusion and masked autoencoder with perceiver}
\label{sec:method}
In this section, we introduce our \textbf{d}iffusion \textbf{a}uto\textbf{e}ncoder with \textbf{p}erceiver (daep) and the corresponding \textbf{m}asked \textbf{a}uto\textbf{e}ncoder with \textbf{p}erceiver (maep)\footnote{Code for both available \href{\codeloc}{here.}}. A unimodal daep has three components: a tokenizer, an encoder, and a diffusion decoder. The corresponding maep has a tokenizer, an encoder, and a direct decoder.

\textbf{Tokenizers.}
We represent raw data as a sequence of tokens in the model dimension. Data are treated as a collection of measurements at specific locations with accompanying metadata. Formally, we define $(\bm{v}, \bm{s}, \bm{m})$, where $\bm{v}$ denotes measurement values (e.g., the flux of an astrophysical source), $\bm{s}$ provides positional information (e.g., wavelength, time, or photometric filter), and $\bm{m}$ encodes observational metadata (e.g., which telescope was used, observation time of spectra). We adapt the perceiver strategy \citep{jaegle2021perceiver} by linearly projecting $\bm{v}$, using fixed sinusoidal embeddings followed by a small MLP for continuous parts of $\bm{s}$ (e.g., time), inspired by \citet{peebles2023scalable}; and categorical embeddings for discrete parts (e.g., photometric filters). We concatenate value and positional embeddings and project them to the model dimension. Metadata are represented as extra tokens and appended to the sequence. 

\begin{figure}[htp]
\centering
\includegraphics[width=0.8\linewidth]{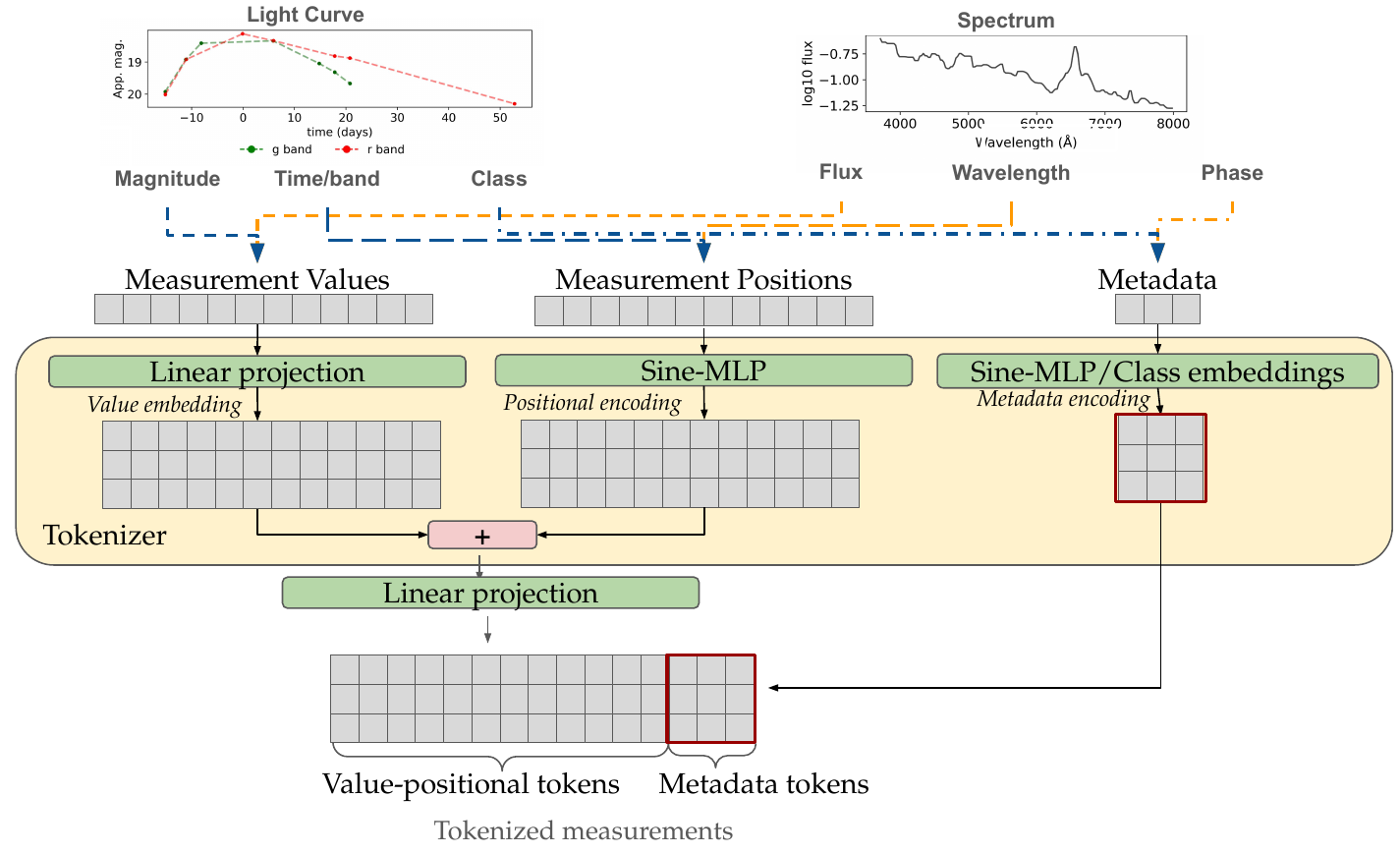}
\vspace{-0.4cm}
\caption{Schematic of tokenizers for general irregularly measured sequences.}
\label{fig:generaltokenizers}
\end{figure}

\begin{figure}[htp]
\centering
\includegraphics[width=0.9\linewidth]{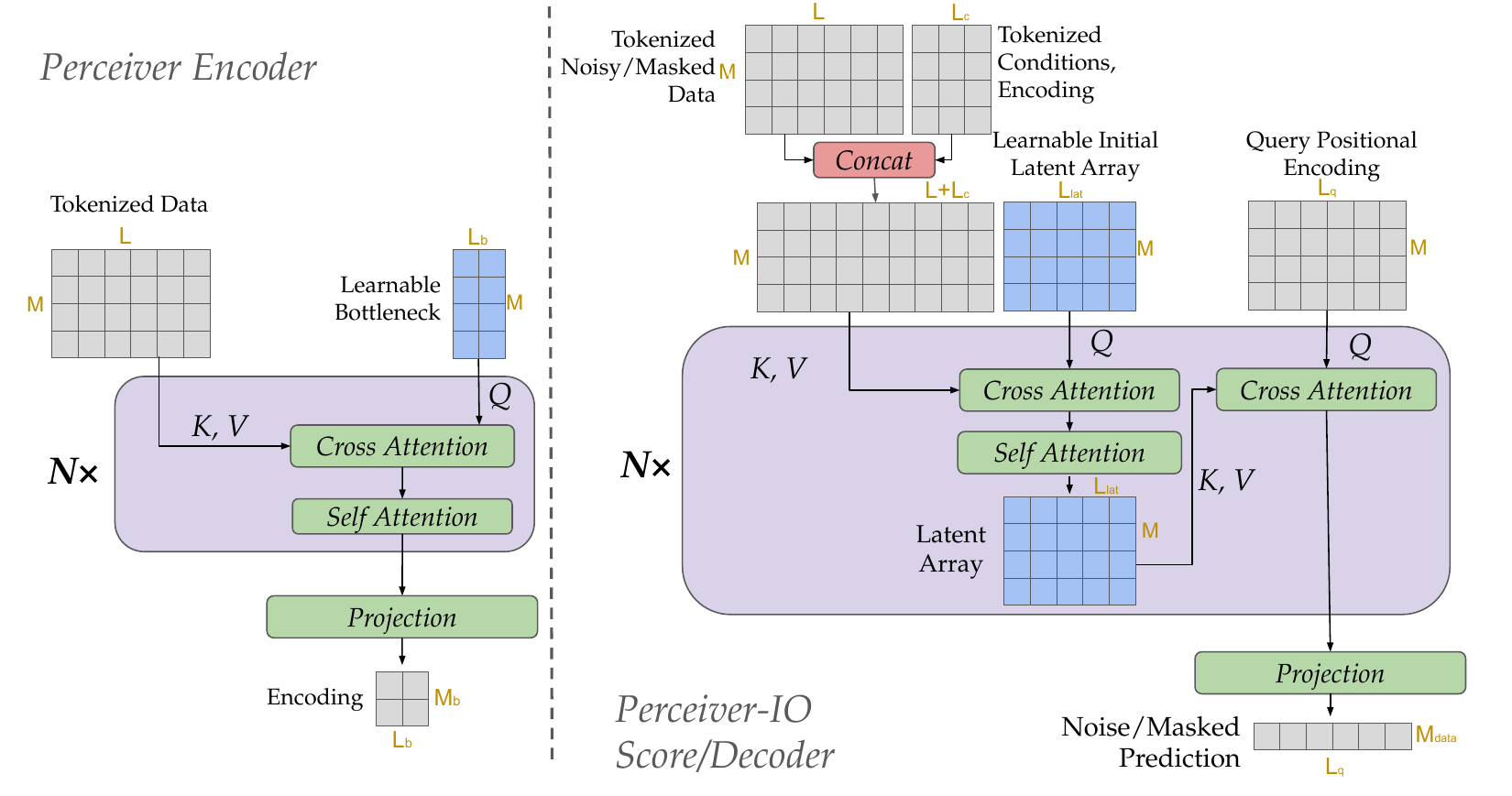}
\vspace{-0.5cm}
\caption{Schematic of the perceiver encoder and perceiver-IO score/decoder model used in daep and mae.}
\label{fig:daep}
\end{figure}

\textbf{Unimodal encoders for both diffusion and masked decoders.}
We use perceiver encoders \citep{jaegle2021perceiver} to map token sequences into compact bottleneck representations, denoted $\text{Enc}_{\theta}$. Input tokens act as Keys and Values in cross-attention, while bottleneck representations serve as Queries. Self-attention is applied only among bottleneck sequences. We repeat these perceiver blocks several times, optionally sharing weights. This design handles variable-length sequences with linear cost in sequence length, making it efficient for processing long and irregular data. Finally, we project bottleneck sequences from the model dimension to a fixed bottleneck dimension. We illustrate the encoder in \cref{fig:daep}. Since perceivers do not require fixed-length input, they can process both masked inputs for MAE training and full data for daep training.

\textbf{Perceiver-IO–based decoder.}
Our diffusion decoder builds on diffusion transformers \citep{peebles2023scalable}, particularly cross-attention conditioning. Diffusion time is encoded with fixed sinusoidal embeddings passed through an MLP, as in \citet{peebles2023scalable}, and concatenated with the conditioning representation for the score model. The score model $\epsilon_{\theta}(x_t, z, t)$, which predicts added noise, is a perceiver-IO: noisy data are tokenized, concatenated with conditioning tokens, and used as Keys and Values in cross-attention. A latent sequence serves as Queries with self-attention, then acts as Keys and Values in a second cross-attention stage with positional information as Queries. We repeat these blocks, optionally sharing weights. The schematic is shown in \cref{fig:daep}. While \citet{jaegle2021perceiverio} recommend latent lengths of 128–512, this may exceed the input sequence length in some tasks. In such cases, we use a single-stage perceiver decoder without a latent sequence, directly connecting noisy tokens to noise prediction through cross-attention.

Because the perceiver-IO architecture is agnostic to query length, the same decoder can be used for the masked autoencoder task as $\text{Dec}_{\theta}(x_{\text{umsk}}, z)$. In this case, we input the tokenized full data sequence with masked locations replaced by a learnable mask token, concatenate with the conditioning $z$ (but no diffusion time $t$), and query using positional information from masked locations only.

\textbf{Training and sampling for daep.}
We train with the score-matching loss $||\epsilon_{\theta}(x_t, \text{Enc}_{\theta}(x_0), t)-\epsilon_t||_2^2$ from DDPM \citep{ho2020denoising}, using 1,000 denoising steps. At inference time, we adopt deterministic DDIM \citep{song2020denoising} for faster sampling with 200 steps. Similar to \citet{preechakul2022diffusion}, our model is not inherently generative, since it requires the bottleneck representation of the input data for conditioned decoding. However, following \citet{preechakul2022diffusion} and \citet{wehenkel2021diffusion}, we can train another DDIM to sample from the bottleneck distribution, enabling prior generation.

\textbf{Training the masked autoencoder.}
We train with a masked reconstruction loss $||x_{\text{msk}}-\text{Dec}_{\theta}(x_{\text{umsk}}, \text{Enc}_{\theta}(x_{\text{umsk}}))||_2^2$. In experiments, we train two variants with different masking ratios of the input: mae-75\% masks 75\% of measurement values, and mae-30\% masks 30\% of measurement values during training. For reconstruction tasks, we provide 10\% of the tokens as unmasked to ensure a relatively fairer comparison with daep and VAEs that does not have access to raw measurements. 

\textbf{VAE baselines.}
For benchmarking, we train a $\beta$-VAE ($\beta=0.1$) with the same perceiver encoder and decoder as both daep and mae. The benchmark models are trained with a weighted sum of the KL-divergence and the L2 reconstruction loss.

\section{Unimodal experiments}
\label{sec:unimodal}
\subsection{High-resolution spectra of variable stars.}

\textbf{Data source.}
We used data from v2.0 DR9 of the Large Sky Area Multi-Object Fiber Spectroscopic Telescope \citep[LAMOST;][]{cui2012large}, specifically the dataset consolidated by \citet{rizhko2025astrom3}. The dataset contains spectra of variable stars, with an average of $\sim$2,500 flux measurements per spectrum and a maximum of $\sim$4,000. In total, we use 17,063 spectra for training and 2,225 for testing. Full architectural details are provided in \cref{app:lamost}.

\begin{figure}[htp]
\centering
\includegraphics[width=1\linewidth]{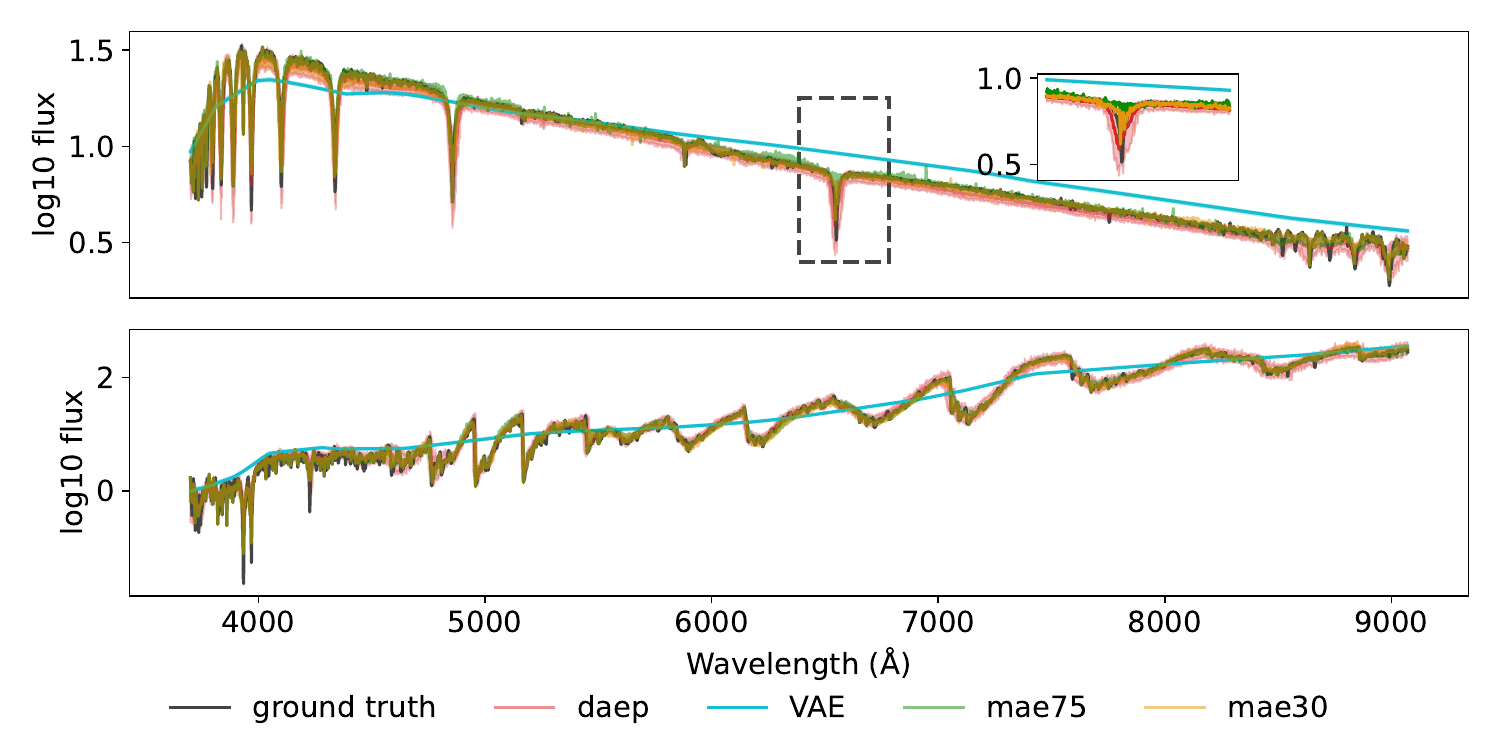}
\vspace{-0.5cm}
\caption{Two example reconstructions of variable star spectra. daep captures finer spectroscopic features, while the VAE mainly reproduces the continuum, likely due to posterior collapse. Both daep and maep successfully capture small-scale spectral features.}
\label{fig:lamostspec}
\end{figure}

\begin{figure}[htp]
\centering
\includegraphics[width=\linewidth]{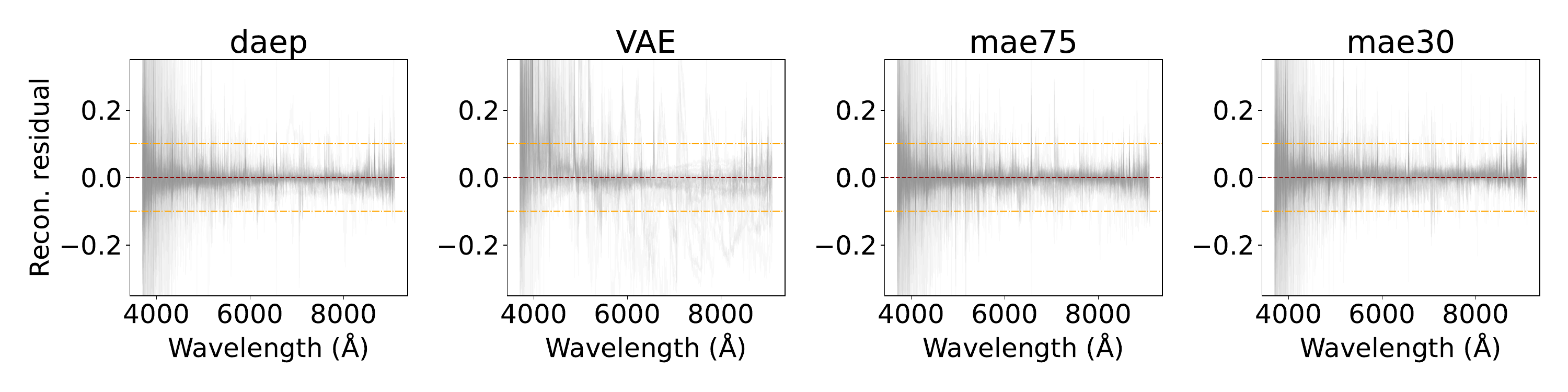}
\vspace{-0.7cm}
\caption{Residuals for test spectra in the LAMOST dataset. Both daep and maep show smaller residuals than VAE. daep also preserves higher-resolution structure, with fewer residual spikes (e.g., near 7000 Å) compared to maep. The red line marks 0 residual while the orange line marks a residual of $\pm$0.1.}
\label{fig:lamostresidual}
\end{figure}

\textbf{Reconstruction.}
In this task, we let the model reconstruct the observed data. For maep models, we encode the full observed data and provide 10\% of unmasked tokens during decoding. We show two enlarged test examples in \cref{fig:lamostspec}, with additional examples in \cref{fig:astrom3more}. Quantifying residual distributions as a function of wavelength is non-trivial because spectra are not aligned on a uniform grid. Instead, we plot all test residuals in \cref{fig:lamostresidual}, with summary metrics in \cref{tab:lamostmetrics}. Both daep and maep achieve superior reconstruction than the baseline $\beta$-VAE, with daep showing fewer residuals in lower-wavelength regions and capturing finer spectral features. Interestingly, for these long sequences, VAEs—even with small $\beta$ values—tend to reproduce only the low-frequency stellar continuum, while daep and maep recover higher-frequency structure in the data.

\textbf{Downstream classification task.}
We classify variable stars into ten classes using linear probing on 30\% of the test set. For MAE, we unmask all measurements during probing. Accuracy and F$_1$ scores are reported in \cref{tab:lamostmetrics}. Both daep and maep outperform VAE, with daep achieving the highest F$_1$ score, indicating stronger representation learning.

\begin{table}[htp]
\centering
\begin{tabular}{lccc}
\hline
Method & Abs. reconstruction error $\downarrow$ & Linear probing Accu. $\uparrow$ & Linear probing F$_1$ $\uparrow$ \\
\hline
daep & \underline{0.038 (0.034)} & \textbf{0.57 (0.003)} & \textbf{0.25 (0.003)} \\
VAE & 0.076 (0.075) & 0.54 (0.002) & 0.24 (0.004) \\
mae-75\% & \underline{0.036 (0.034)} & 0.55 (0.002) & 0.23 (0.002) \\
mae-30\% & \textbf{0.036 (0.029)} & 0.56 (0.002) & 0.23 (0.003) \\
\hline
\end{tabular}
\caption{Reconstruction and downstream linear probing metrics on LAMOST spectra. Best-performing models are boldfaced; underlined results indicate models whose 1,std interval overlaps with the best mean.}
\label{tab:lamostmetrics}
\end{table}

\subsection{Low-resolution spectra of supernovae.}

\textbf{Data source.}
Next, we use data from the Zwicky Transient Facility Bright Transient Survey \citep[ZTFBTS;][]{bellm2018zwicky}. The spectra were obtained from multiple different facilities and corrected for redshift, resulting in irregular grids in measurement position (wavelength). We encoded the spectra into four latent tokens of dimension four.

\begin{figure}[htp]
\centering
\includegraphics[width=\linewidth]{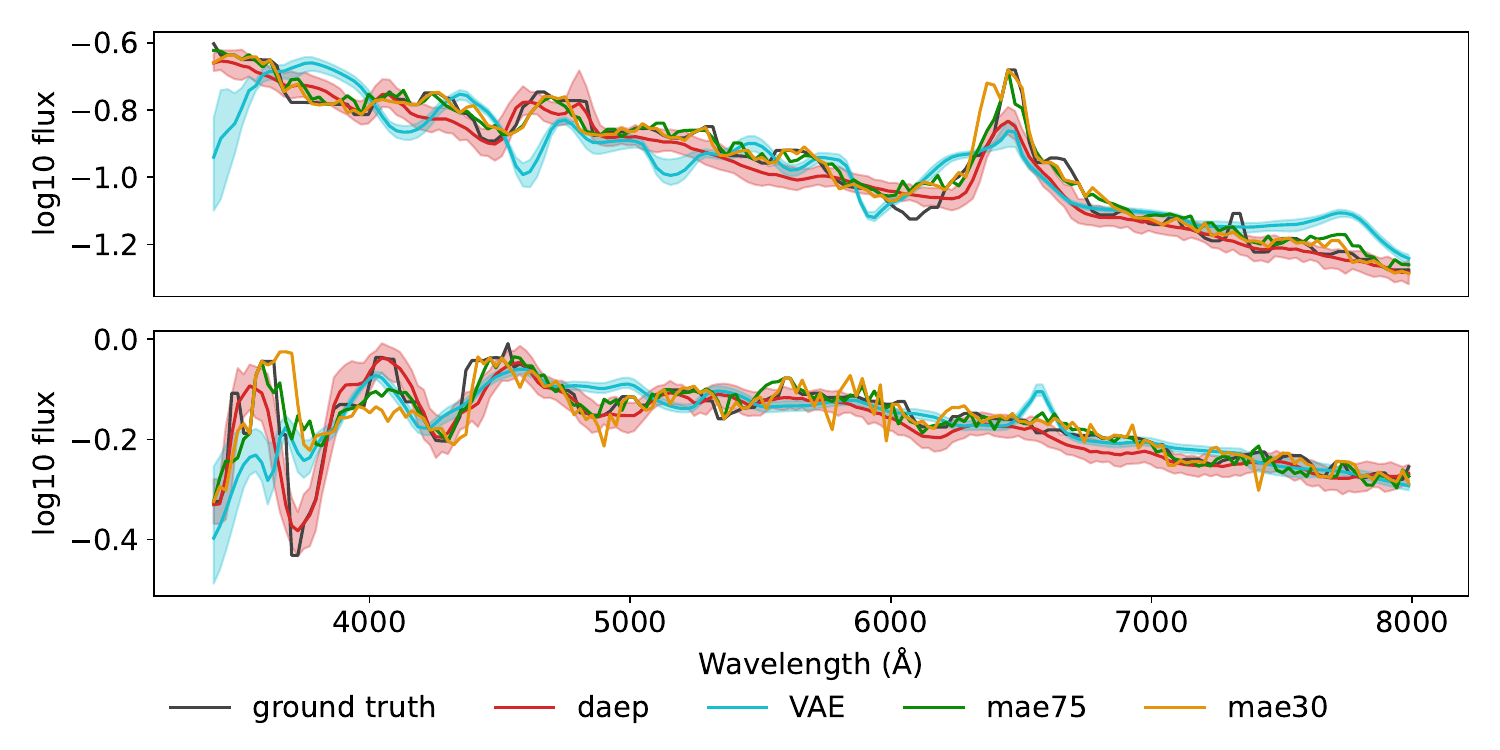}
\vspace{-0.6cm}
\caption{Reconstruction of SEDM spectra for ZTF supernovae from four latent tokens of dimension four using daep and baseline models.}
\label{fig:ztfspectra}
\end{figure}

\begin{figure}[htp]
\centering
\includegraphics[width=\linewidth]{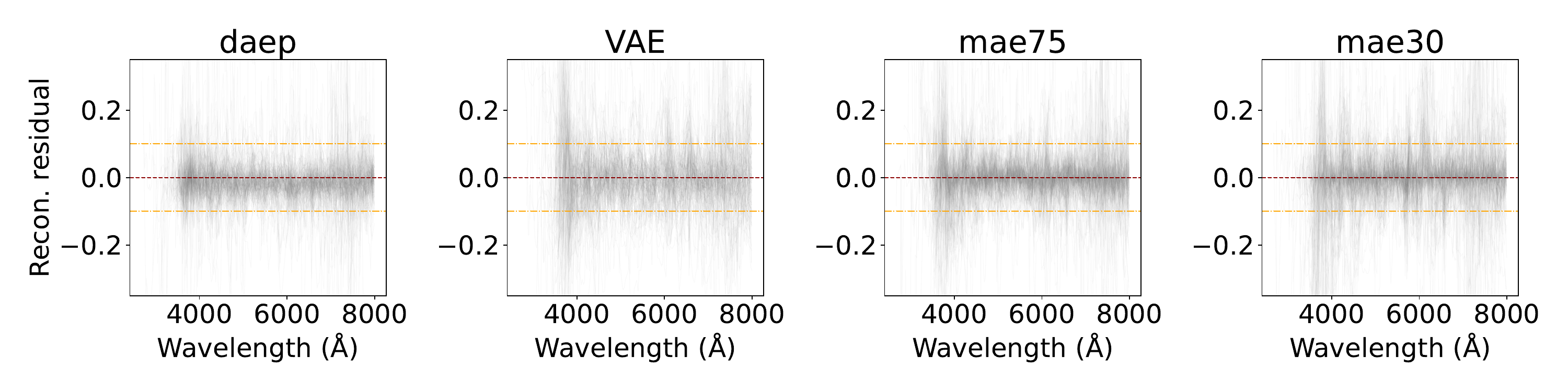}
\vspace{-0.6cm}
\caption{Residuals for test spectra from the ZTF BTS sample. daep achieves consistently lower reconstruction residuals across the considered wavelength range compared to the VAE and maep baselines. Red line marked 0 residual while orange line marked at $\pm$0.1.}
\label{fig:ztfspectraresi}
\end{figure}

\textbf{Reconstruction.}
As before, we let the model reconstruct the observed data. For maep models, we encode the full observed data and provide 10\% of unmasked tokens during decoding. We provide two test examples in \cref{fig:ztfspectra}, with additional examples in \cref{fig:ztfspectramore}. The population-level reconstruction error is visualized in \cref{fig:ztfspectraresi}. For this task, daep produces smaller residuals than both VAE and maep models across the full wavelength range, consistent with the metrics in \cref{tab:metricZTFspectra}. This may be due to the diffusion decoder’s ability to better handle noise and data artifacts, which contaminate the low-resolution ZTF spectra more severely than the LAMOST spectra.

\textbf{Downstream classification task.}
Next, we perform linear probing on a three-way  classification task to distinguish type Ia, type Ib/c, and other supernovae. For MAE, we unmask all measurements during probing. Results are reported in \cref{tab:metricZTFspectra}. daep achieves both the highest classification accuracy and the highest F$_1$ score. 

\begin{table}[htp]
\centering
\begin{tabular}{lccc}
\hline
Method & Abs. reconstruction error $\downarrow$ & Linear probing Accu. $\uparrow$ & Linear probing F$_1$ $\uparrow$ \\
\hline
daep & \textbf{0.040 (0.028)} & \textbf{0.82 (0.009)} & \textbf{0.45 (0.020)} \\
VAE & 0.070 (0.047) & 0.81 (0.002) & 0.40 (0.009) \\
mae-75\% & \underline{0.047 (0.026)} & 0.79 (0.002) & 0.34 (0.010) \\
mae-30\% & \underline{0.066 (0.032)} & 0.79 (0.001) & 0.30 (0.003) \\
\hline
\end{tabular}
\caption{Reconstruction and downstream classification metrics for ZTF BTS spectra. Reconstruction metrics are averaged over events; classification metrics are averaged over 10 probe/evaluation splits. Best models are boldfaced; underlined results overlap with the best mean within 1 standard deviation.%\textcolor{red}{this is minor but because of class imbalance folks evaluate both macro and micro-averaged F1 and accuracy in SN classification. Would be useful to compare here, since the dataset is mostly Ia.}
}
\label{tab:metricZTFspectra}
\end{table}

\subsection{Photometry of supernovae.}

\textbf{Data source.}
We used photometry from the ZTF BTS \citep{bellm2018zwicky}. Supernova flux was measured in two photometric filters, or ``bands" — green (g) and red (r) — along with spectra, all sampled irregularly in time. We encoded this photometry (collectively denoted the supernova ``light curve") into a two-token sequence of dimension two.

\begin{figure}[htp]
\centering
\includegraphics[width=0.9\linewidth]{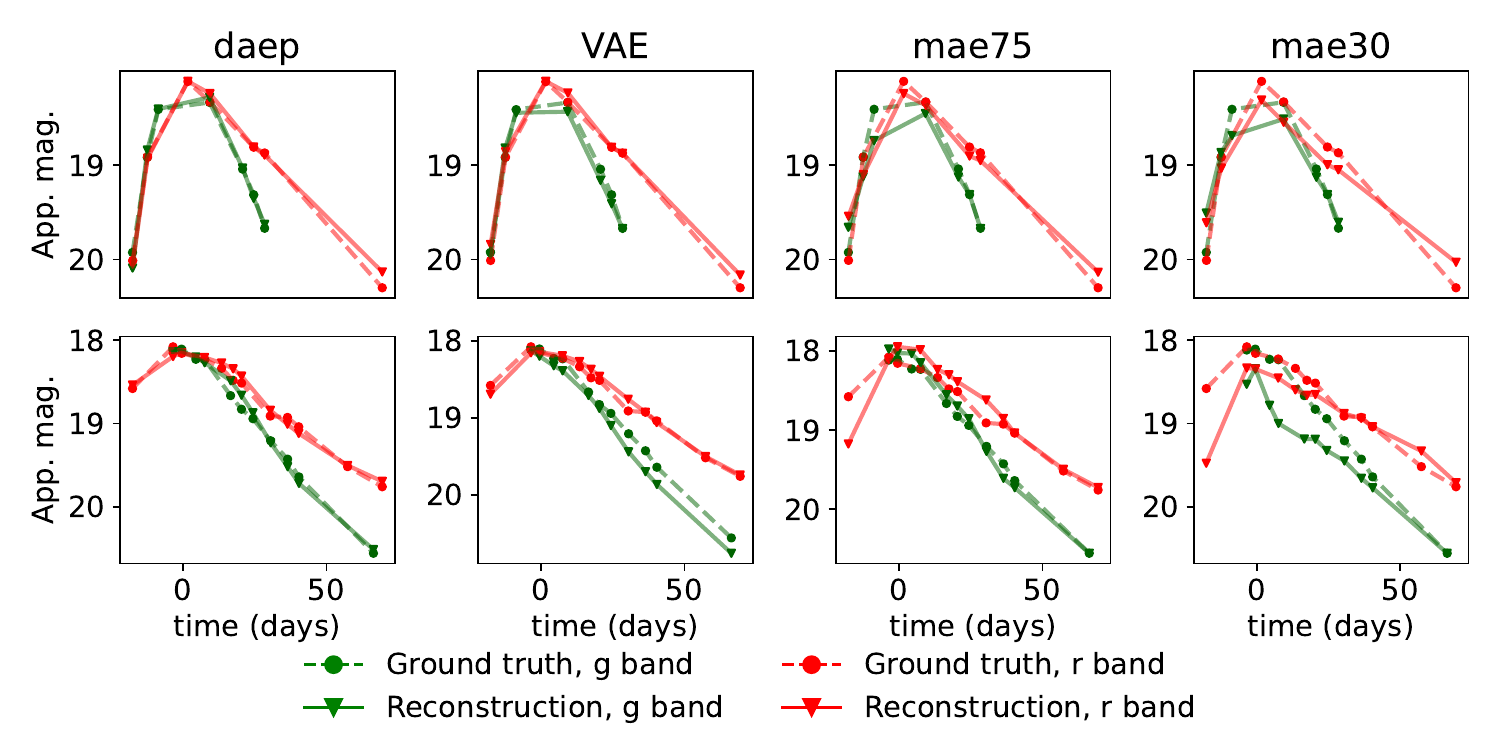}
\vspace{-0.3cm}
\caption{Light curve reconstruction from two latent tokens of dimension two using daep and baselines. (App. mag. stand for apparent magnitude).%\textcolor{red}{vertical axis: Mag --> Apparent Mag}
}
\label{fig:lc}
\end{figure}

\begin{figure}[htp]
\centering
\includegraphics[width=\linewidth]{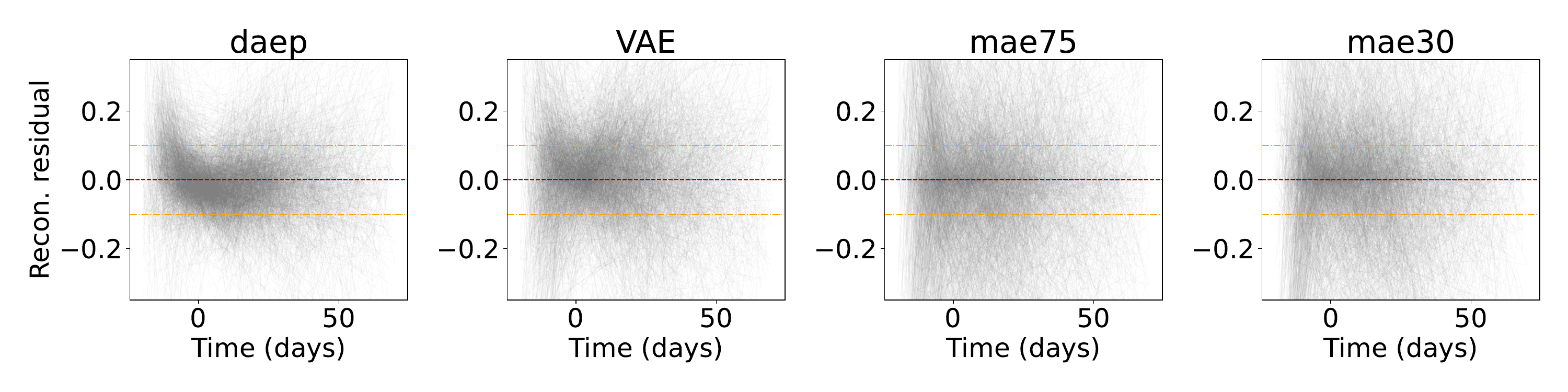}
\vspace{-0.8cm}
\caption{Residuals for test light curves from ZTF photometry. daep achieves smaller residuals overall than both VAE and MAE, but systematically overestimates the brightness before peak light. Red line marked 0 residual while orange line marked at $\pm$0.1.}
\label{fig:ztflcresidual}
\end{figure}

\textbf{Reconstruction.}
In this task, we let the model reconstruct the observed data. For MAE-based models, we again provide 10\% unmasked tokens during decoding. Example reconstructions are shown in \cref{fig:lc}, with more in \cref{fig:ztflcvae}. Residuals are plotted in \cref{fig:ztflcresidual}. daep achieves more accurate reconstructions than VAE, as also reflected in \cref{tab:metricZTFphotometry}. Interestingly, daep tends to overestimate brightness before peak ($<0$ days), an effect not observed in other models, but still achieves the highest-fidelity reconstruction.

\textbf{Downstream classification.}
We perform linear probing on a three-class task to classify supernovae as type Ia, type Ib/c, or other. For MAE, we use full measurements during encoding; other models use the latent. Results are shown in \cref{tab:metricZTFphotometry}. In this task, maep achieves the best performance, likely because it better encodes general trends, while photometry contains fewer high-frequency features for daep to capture.

\begin{table}[htp]
\centering
\begin{tabular}{lccc}
\hline
Method & Abs. reconstruction error $\downarrow$ & Linear probing Accu. $\uparrow$ & Linear probing F$_1$ $\uparrow$ \\
\hline
daep & \textbf{0.078 (0.038)} & 0.821 (0.023) & 0.414 (0.061) \\
VAE & \underline{0.110 (0.064)} & 0.773 (0.040) & 0.413 (0.052) \\
mae-75\% & \underline{0.102 (0.083)} & \underline{0.882 (0.014)} & 0.550 (0.024) \\
mae-30\% & \underline{0.091 (0.066)} & \textbf{0.890 (0.003)} & \textbf{0.630 (0.044)} \\
\hline
\end{tabular}
\caption{Reconstruction and downstream classification metrics for ZTFBTS photometry. Reconstruction metrics are averaged over events; classification metrics are averaged over 10 probe/evaluation splits. Best models are boldfaced; underlined results overlap with the best mean within 1 std. }
\label{tab:metricZTFphotometry}
\end{table}

\section{Multimodal with modality dropping}
\label{sec:multimodal}
\begin{figure}[htp]
\centering
\includegraphics[width=0.6\linewidth]{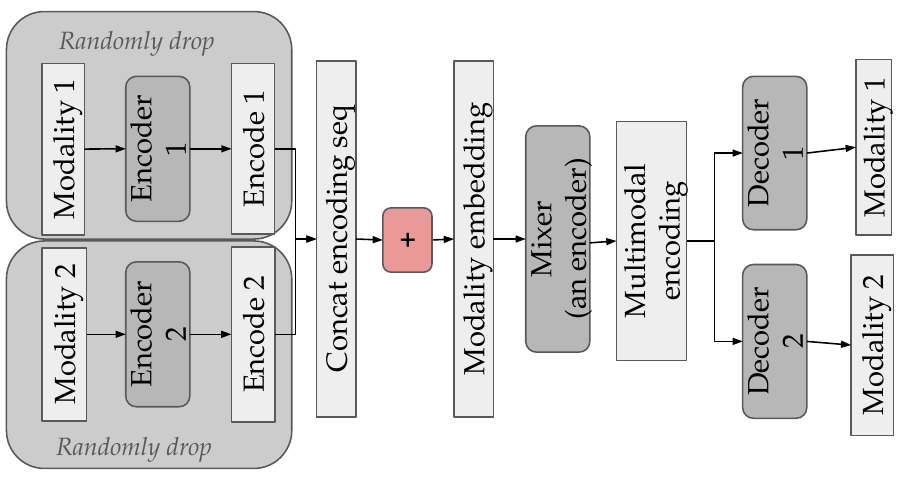}
\caption{Late mixing and modality dropping for the multimodal daep model.}
\label{fig:latefusion}
\end{figure}

\textbf{Modality mixing and training for multimodal data.}
To learn joint representations from multiple modalities, we use a late-mixing strategy. The goal is to have a latent representation that summarize all modalities in hand. This cannot directly be done with e.g., mixture of expert VAE or contrastive learning where each modality has own encoding. Each modality is first encoded with a perceiver encoder, a learnable modality embedding is then added to all tokens from that modality, and concatenated along the sequence dimension. A second perceiver encoder then acts as a “mixer” to produce a single compact bottleneck sequence. Because the perceiver encoder does not require fixed-length input, we employ modality dropping \citep{neverova2015moddrop, liu2022moddrop++} during training so the multimodal model can accommodate missing modalities. All modalities are decoded using modality-specific diffusion decoders. A schematic of this architecture with two modalities is shown in \cref{fig:latefusion}.

\textbf{Simulated supernova spectra and photometry.}
\begin{table}[htp]
    \centering
    \begin{tabular}{lccccc}
    \hline
        Method & MSE -10 days & MSE 0 days & MSE 10 days & MSE 20 days & MSE 30 days \\
    \hline
    mmVAE    & \textbf{0.080 (0.05)} & \textbf{0.006 (0.005)} & \textbf{0.005 (0.005)} & \textbf{0.006 (0.009)} & \underline{0.111 (0.021)} \\
    daep    & \underline{0.163 (0.10)} & \underline{0.017 (0.015)} & \underline{0.012 (0.011)} & \underline{0.011 (0.013)} & \textbf{0.019 (0.027)} \\
    contrastive    & 0.549 (0.43) & 0.073 (0.032) & 0.105 (0.050) & 0.090 (0.048) & 0.110 (0.057) \\

    \hline     
    \end{tabular}
    \caption{Performance of the cross-modality inference task from photometry to spectra on the simulated dataset. We boldface the best-performing model and underline those whose 1 std interval contains the best mean. Our method performs similarly to mmVAE and outperforms contrastive search.}
    \label{tab:goldsteintable}
\end{table}

\begin{figure}[htp]
    \centering
    \includegraphics[width=\linewidth]{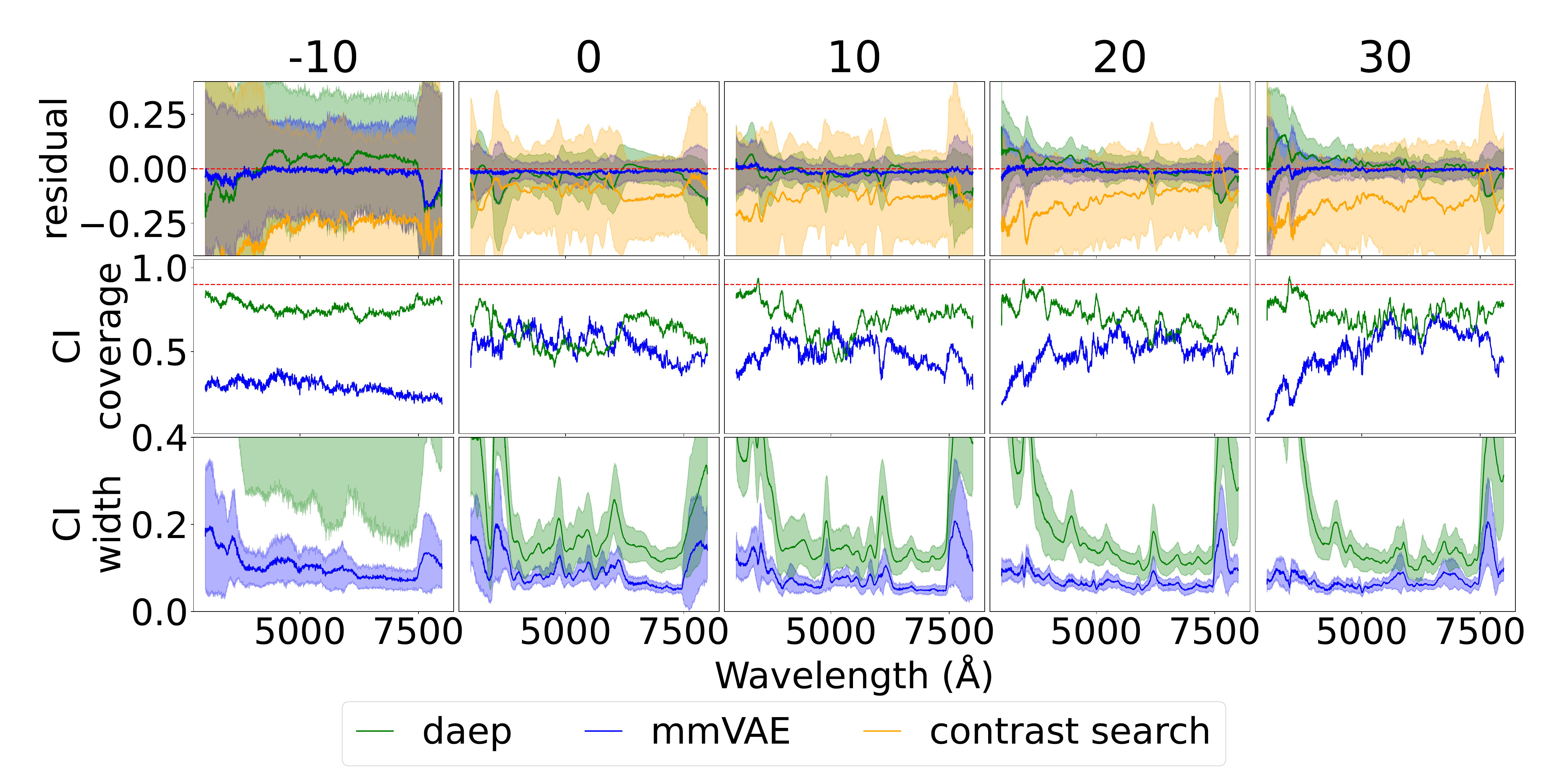}
    \vspace{-0.5cm}
    \caption{Performance of cross-modality inference with daep, mmVAE, and contrastive nearest-neighbor search. Our method achieves similar reconstruction performance to mmVAE, and superior to contrastive search. Our method's CI widths also exhibit better coverage than mmVAE at all phases relative to explosion, and are near-optimal (red line).}
    \label{fig:goldsteinfig}
\end{figure}
We use the 5,000 simulated type~Ia supernovae from the radiative transfer models in \citet{goldstein2018evidence}. Following the approach in \citet{shen2025mixture}, we post-process the spectral energy distributions of each event to simulate idealized photometry with six filters from the Vera C. Rubin Observatory LSST in Wide-Fast-Deep mode \citep{2019Ivezic_lsst}. Each light curve is paired with five spectra taken at $-10, 0, 5, 10, 20,$ and $30$ days after peak brightness. We focus on the cross-modality inference task of reconstructing spectra from photometry. This task is common one in time-domain astrophysics: photometry is inexpensive to obtain while spectra require significantly higher integration times, and spectroscopic datasets are therefore significantly more sparse. Since MAEs cannot handle completely missing modalities, we benchmark against (1) a mixture-of-experts VAE and (2) the contrastive-learning-based nearest-neighbor search from \citet{shen2025mixture}.

We evaluate model performance using residuals, confidence interval (CI) coverage, and CI width as a function of wavelength and observation time, as shown in \cref{fig:goldsteinfig} and \cref{tab:goldsteintable}. Our method performs similarly to mmVAE, with slightly better coverage. Both substantially outperform contrastive search. This is likely because photometry contains less information than spectra, leading to modality collapse in weaker baselines (a similar conclusion is made by the authors of \citealt{zhang2024maven} using a comparable dataset).

\section{Discussion}
\label{sec:discussion}
In this work, we have presented an architecture for reconstruction-based SSL on multivariate sequential datasets. Although we validate daep primarily on astrophysical datasets, the architecture is domain-agnostic and directly applicable to other irregular, multimodal domains such as healthcare, finance, and sensor networks. The combination of perceiver tokenization and diffusion decoding enables scalable self-supervised learning on data that existing SSL methods cannot readily accommodate.

Our comparisons with perceiver-based MAEs reveal that decoder context plays a critical role: MAEs benefit from access to unmasked tokens, yet daep performs comparably or better without such context, and outperforms MAEs without decoder access. This highlights daep’s ability to compress high-frequency information into a latent bottleneck. We also find that daep reconstructions more faithfully capture high-frequency spectral features, essential for enabling downstream tasks reliant on fine detail (e.g., the identification of short-duration signal anomalies).

While daep is more flexible than existing approaches, diffusion decoding is computationally heavier than masking, and our experiments focus on astronomy. Future work will broaden the evaluation to clinical, financial, and multimodal sensor datasets, and explore hybrid objectives that combine diffusion reconstruction with predictive or contrastive tasks \citep{huang2023contrastive} for better cross modality alignment, as well as using measurement noise aware losses. Beyond representation learning, daep may also serve as a generative model for simulating complex irregular multimodal phenomena and augmenting existing datasets. We plan to explore these extensions in future work.

%\newpage
\bibliography{iclr2025_conference}

%\newpage
\appendix
\section*{Appendix}
%You may include other additional sections here.
\renewcommand{\thefigure}{A\arabic{figure}}
\renewcommand{\thetable}{A\arabic{table}}
\setcounter{figure}{0}
\setcounter{table}{0}

%\section{Prior sampling}
\section{Implementation details}
\label{app:implementation}
%\subsection{Tokenizer details}
%\begin{figure}[htp]
%\centering
%\includegraphics[width=1\linewidth]{Figs/daep-tokenizer.pdf}
%\caption{Tokenizers used in our empirical studies, from left to right: spectra (flux across wavelengths) and light curves (brightness in different colors over time).}
%\label{fig:tokenizer}
%\end{figure}

\subsection{LAMOST model details}
\label{app:lamost}

\textbf{Data preprocessing}
We enforced a 3-$\sigma$ quality cut, i.e., only measurements exceeding 3 times the measurment error are kept for modeling. After all quality cuts, we have 17,063 training stars. We took arcsinh of flux before modeling and after generation we calculate the original flux. Flux and wavelength are standardized to a z-score using the mean and standard deviation of the full training set after calculating arcsinh of flux. 

\textbf{Architectural details}
\begin{table}[htp]
    \centering
    \small
    \begin{tabular}{lccccccc}
    \hline
         model & \makecell{bottleneck\\len} & \makecell{bottleneck\\dim} & \makecell{enc.\\layers} & \makecell{dec.\\layers} & \makecell{model\\dim} & \makecell{\#\\heads} &  \makecell{hidden\\seq len} \\
    \hline
        daep & 4 & 8 & 4 & 4 & 128 & 8 & 256\\
        mae & 4 & 8 & 4 & 2 & 128 & 8 & 256\\
        VAE & 4 & 8 & 4 & 4 & 128 & 8 & 256\\
    \hline
    \end{tabular}
    \caption{Architectural parameters for the model used in LAMOST experiments. MLPs in transformers have dimension twice of the model dimension. }
    \label{tab:hyperparamLAMOST}
\end{table}
\textbf{Training details}
We used a learning rate of $2.5\times10^{-4}$ for both models and trained for 2000 epochs and 200 epochs respectively for daep and VAE, confirming that the training loss converged for both models. We set $\beta=0.1$ for the VAE. 

\subsection{ZTF model details}
\label{app:ztfmodel}

\textbf{Light curve preprocessing}
We first enforced a 3-$\sigma$ cut on measurements, then used a Gaussian process to find the peak time of red band as the 0 phase We align time to be relative to the peak time. We only kept events whose light curves have measurement before and after the peak. 

\textbf{Spectra preprocessing}
We enforce a 3-$\sigma$ quality cut for both spectra and light curve, i.e., only measurements exceeding 3 times the measurement error are kept for modeling. After all cuts, we have 2,934 events left in the training set. We take the base-10 logarithm of the flux before modeling, and after generation we calculate the original flux. We also apply a median filter to filter out noise. Flux and wavelength values are then standardized to a z-score using the mean and standard deviation of the full training set. 

\textbf{Architectural details}
\begin{table}[htp]
    \centering
    \begin{tabular}{lcccccc}
    \hline
         model & bottleneck len & bottleneck dim & enc. layers & dec. layers & model dim & \# heads  \\
    \hline
        daep & 4 & 4 & 4 & 4 & 128 & 8 \\
        mae & 4 & 4 & 4 & 2 & 128 & 8 \\
        VAE & 4 & 4 & 4 & 4 & 128 & 8 \\
    \hline
    \end{tabular}
    \caption{Architectural parameters for the model used in our ZTF supernova spectra experiments. We used a single-stage decoder (skipping the latent sequence) since the sequence is short. MLPs in transformers have dimension twice of the model dimension. }
    \label{tab:hyperparamZTF}
\end{table}

\begin{table}[htp]
    \centering
    \begin{tabular}{lcccccc}
    \hline
         model & bottleneck len & bottleneck dim & enc. layers & dec. layers & model dim & \# heads  \\
    \hline
        daep & 2 & 2 & 4 & 4 & 128 & 4 \\
        mae & 2 & 2 & 4 & 2 & 128 & 4 \\
        VAE & 2 & 2 & 4 & 4 & 128 & 4 \\
    \hline
    \end{tabular}
    \caption{Architectural parameters for the model used in our ZTF light curve experiments. We used a single-stage decoder (skipping the latent sequence) since the sequence is short.}
    \label{tab:hyperparamZTF}
\end{table}

\textbf{Training details}
In contrast from our LAMOST experiment, we augment our data by 5 folds, adding noise to flux measurement and randomly masking measurements due to the small dataset size. We used same learning rate of $2.5\times10^{-4}$ for both models and trained for 2000 epochs and 200 epochs respectively for daep and VAE, confirming that the training loss converged for both models. We set $\beta=0.1$ for the VAE.

\subsection{Multimodal spectra and photometry}
\label{app:goldsteinmodel}
\textbf{Data preprocessing.} We did not perform further processing beyond the steps outlined in \citet{shen2025variational}. 

\textbf{Architectural details}
We have the first stage encoder for both light curves and spectra to have a model dimension of 256, 4 layers, 4 heads, and 64 tokens after encoding. The modality mixer has 4 layers and 4 heads and model dimension 256, and during encoding we allow the concatenated sequence to attend to itself. We encode to a bottleneck sequence of 4 tokens of dimension 4 each. 

\textbf{Training details.}
In each batch, we randomly dropped each data modality with probability 0.2, making sure that at least one modality is retained. We trained with a learning rate $2.5\times10^{-4}$ and for 2000 epochs, confirming convergence of the loss.

%\newpage
\section{Further experimental results}
\subsection{LAMOST spectra}
\begin{figure}[H]
    \centering
    \includegraphics[width=1.\linewidth]{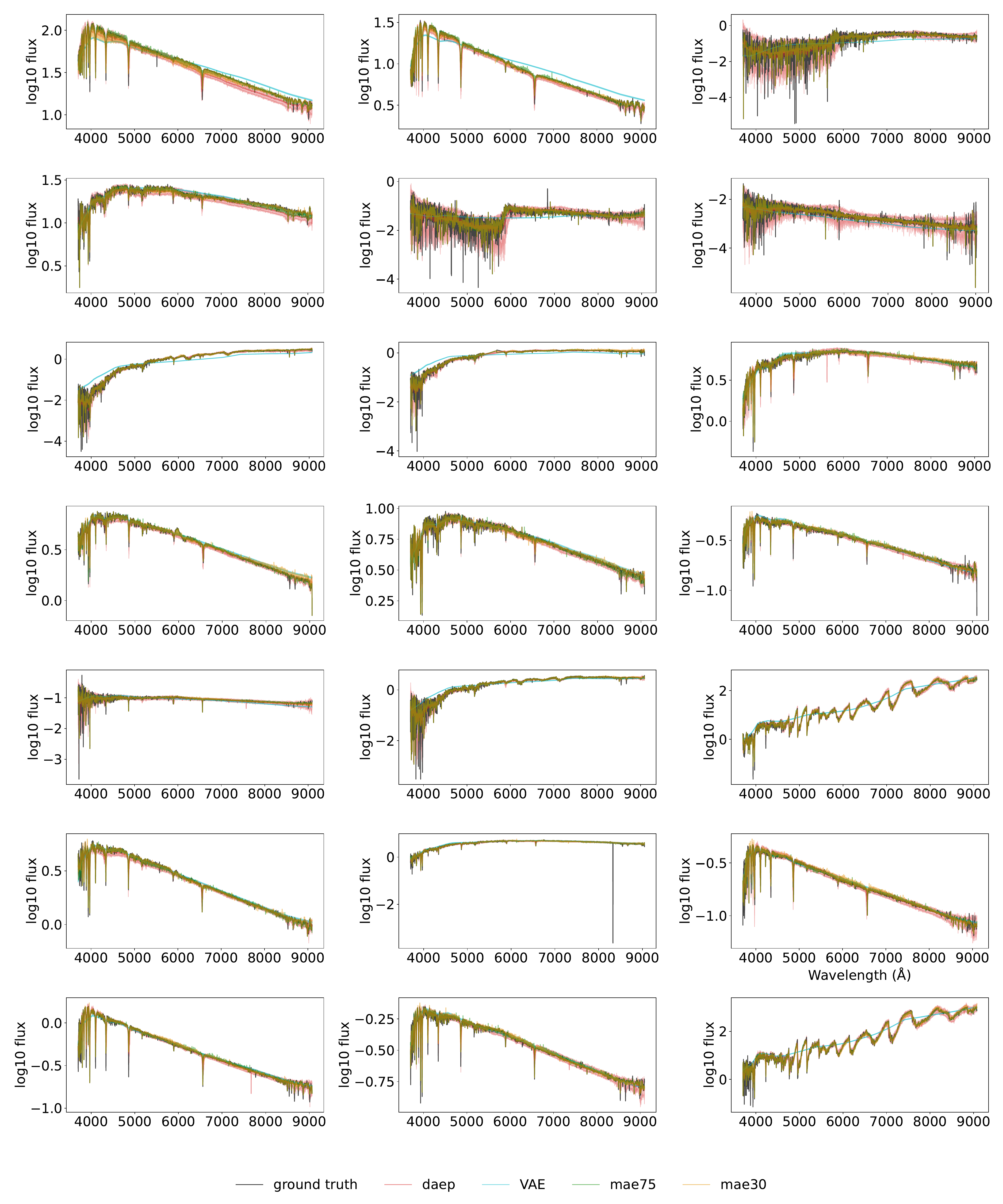}
    \caption{Reconstructions of additional LAMOST variable star spectra. Our method (red) captures more high-frequency absorption features than the VAE baseline with the same-sized bottleneck representation (blue).}
    \label{fig:astrom3more}
\end{figure}

In \cref{fig:astrom3more}, we show additional spectra reconstructions using daep and VAE baselines. Our method consistently captures higher-frequency information details compared to the VAE baseline.  

\begin{figure}[H]
    \centering
    \includegraphics[width = \linewidth]{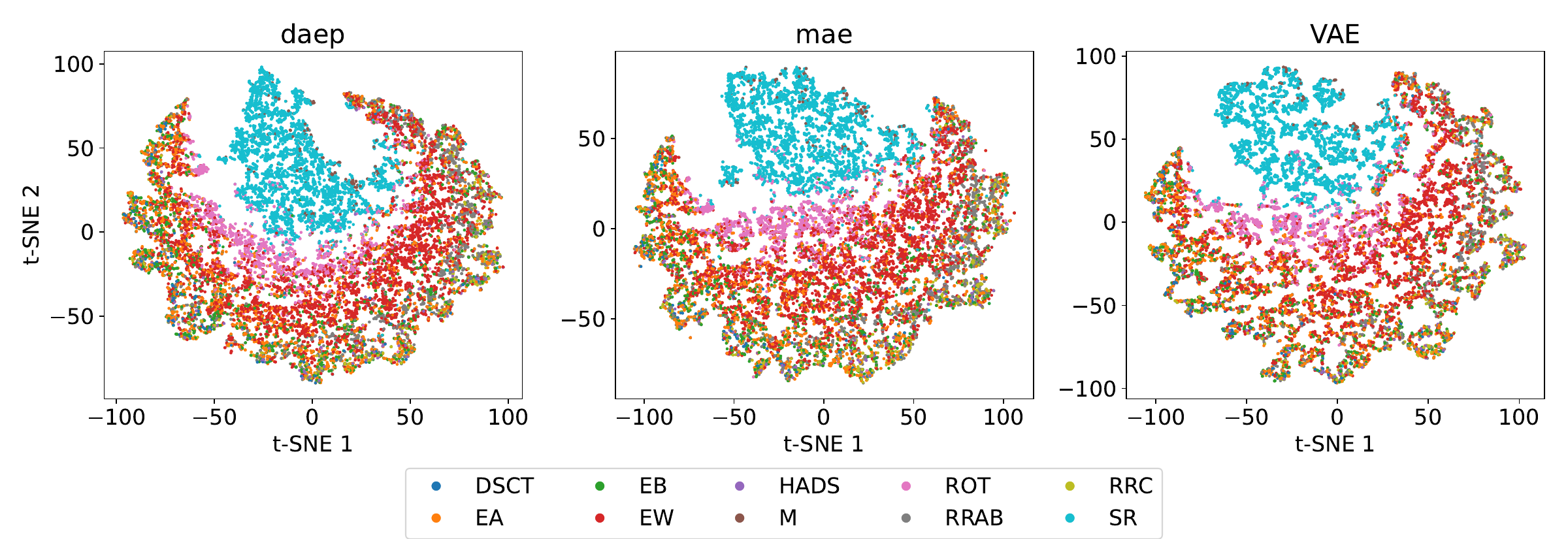}
    \includegraphics[width = \linewidth]{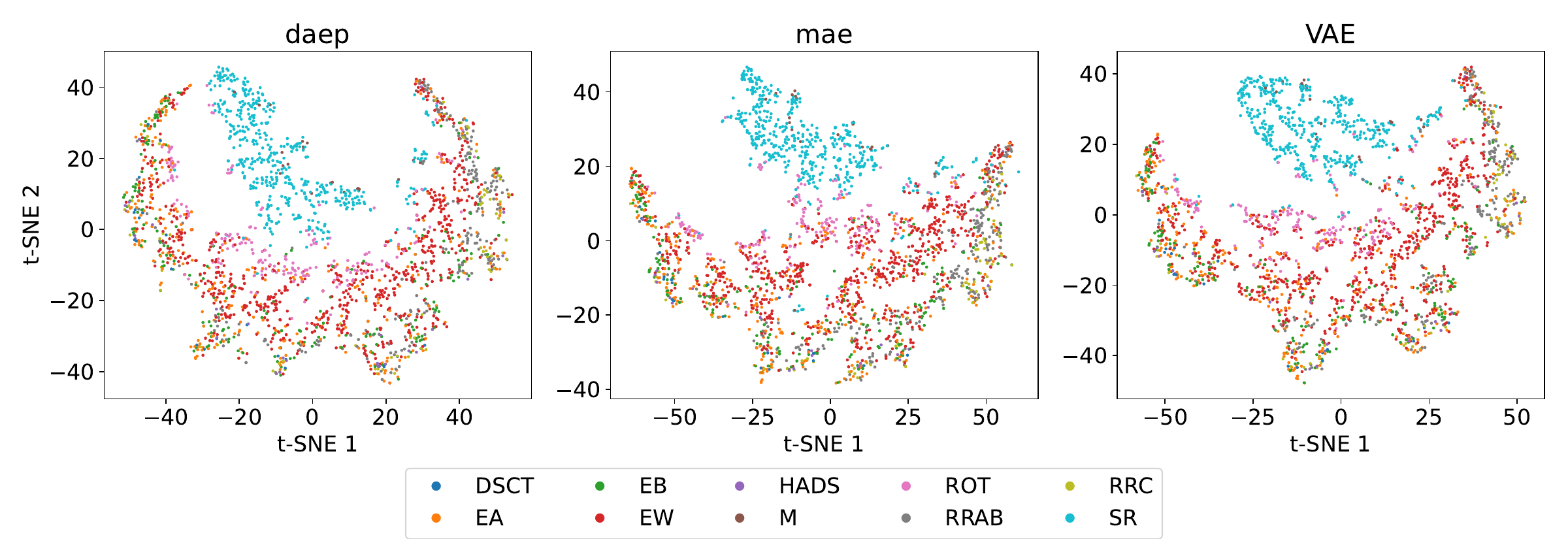}
    \caption{Latent representations of LAMOST spectra from training (upper) and test (lower) sets. The latent space for daep appears more well-regularized compared to a VAE of comparable dimensionality.}
    \label{fig:astrom3specencode}
\end{figure}

In \cref{fig:astrom3specencode}, we compare the latent representations of LAMOST spectra (after t-SNE) from daep, mae and VAE, colored by variable star classification. 

\newpage
\subsection{ZTF spectra}
\label{app:ztfspectrares}
\begin{figure}[h]
    \centering
    \includegraphics[width=1.\linewidth]{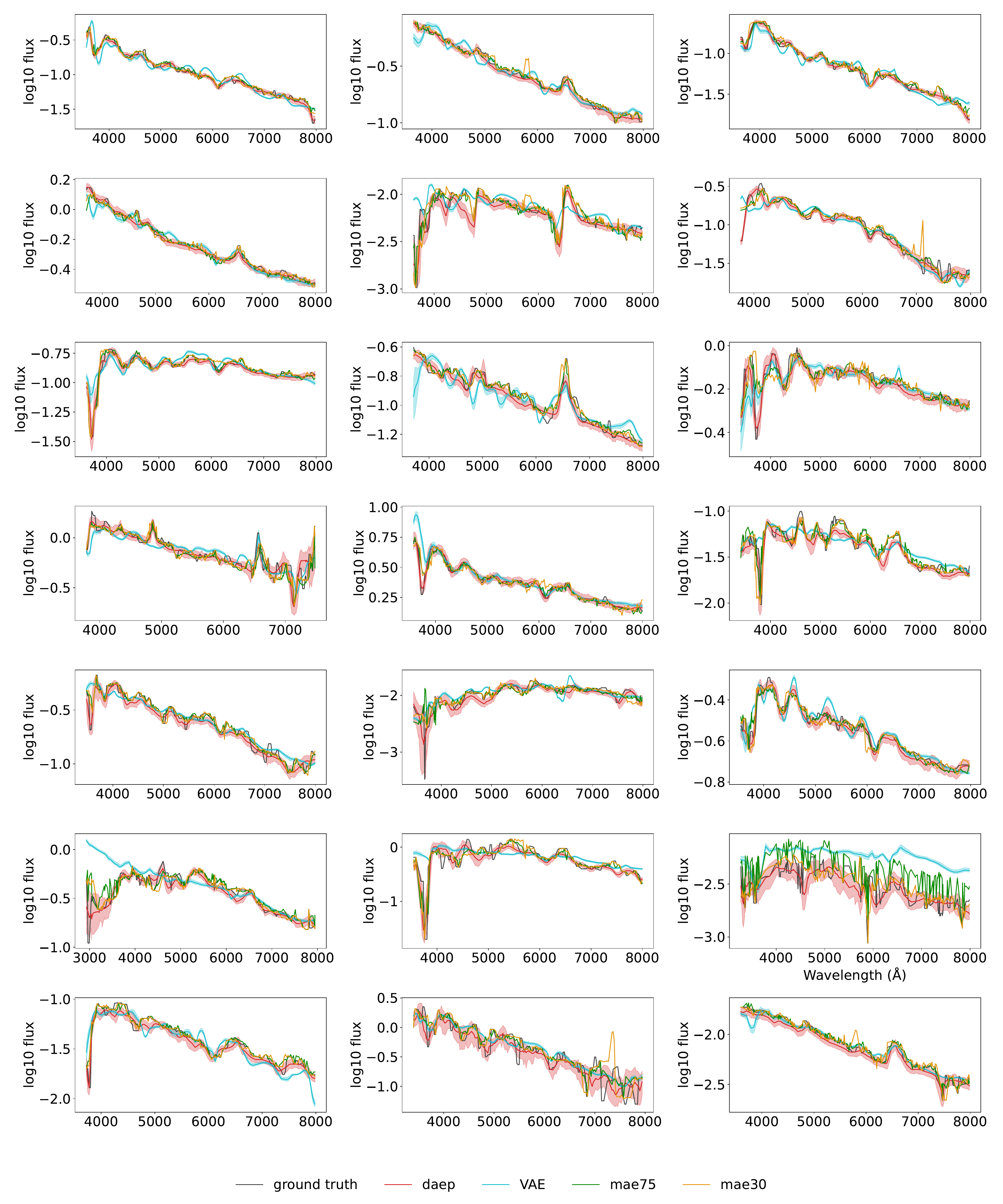}
    \caption{Additional ZTF spectra reconstructions with daep and VAE. Our method captures finer details and maintains better posterior coverage.}
    \label{fig:ztfspectramore}
\end{figure}

In \cref{fig:ztfspectramore}, we show additional spectroscopic reconstructions for ZTF BTS supernovae. Our method captures finer details and produces better-covered posteriors than the VAE baseline.  

\begin{figure}[h]
    \centering
    \includegraphics[width = \linewidth]{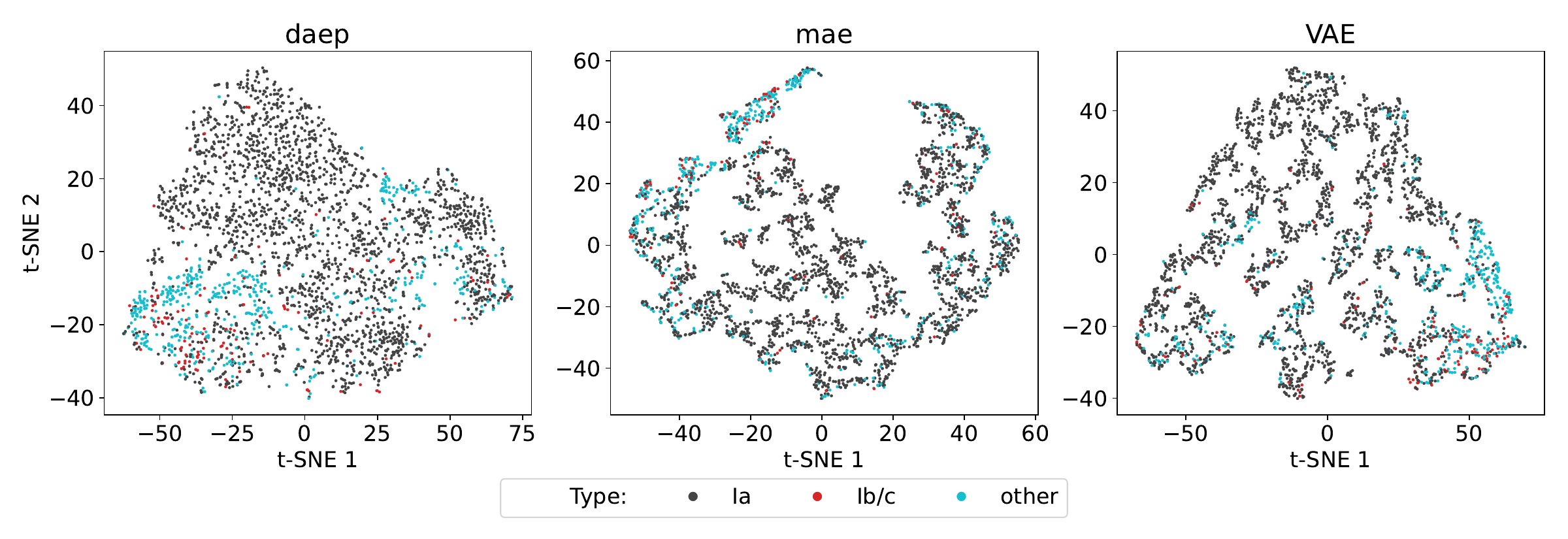}
    \includegraphics[width = \linewidth]{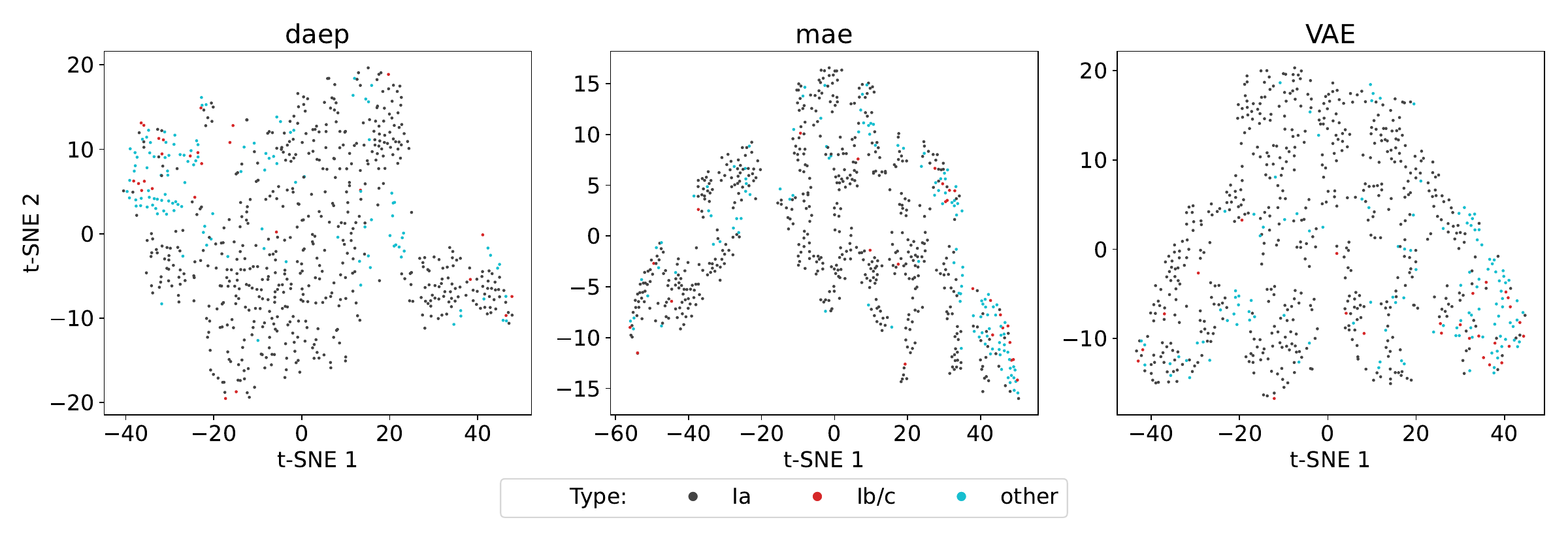}
    \caption{Latent representations of ZTF spectra from training (upper) and test (lower). The latent space for daep appears more well-regularized compared to a VAE of comparable dimensionality.}
    \label{fig:ztfspecencode}
\end{figure}

In \cref{fig:ztfspecencode}, we compare latent representations of the ZTF spectra (after t-SNE) from daep, mae and VAE, colored by event type. Interestingly, the daep latent space appears more continuous than either the MAE or the VAE with $\beta=0.1$.  

\newpage
\subsection{ZTF light curves}
\begin{figure}[htp]
    \centering
    \includegraphics[width=0.85\linewidth]{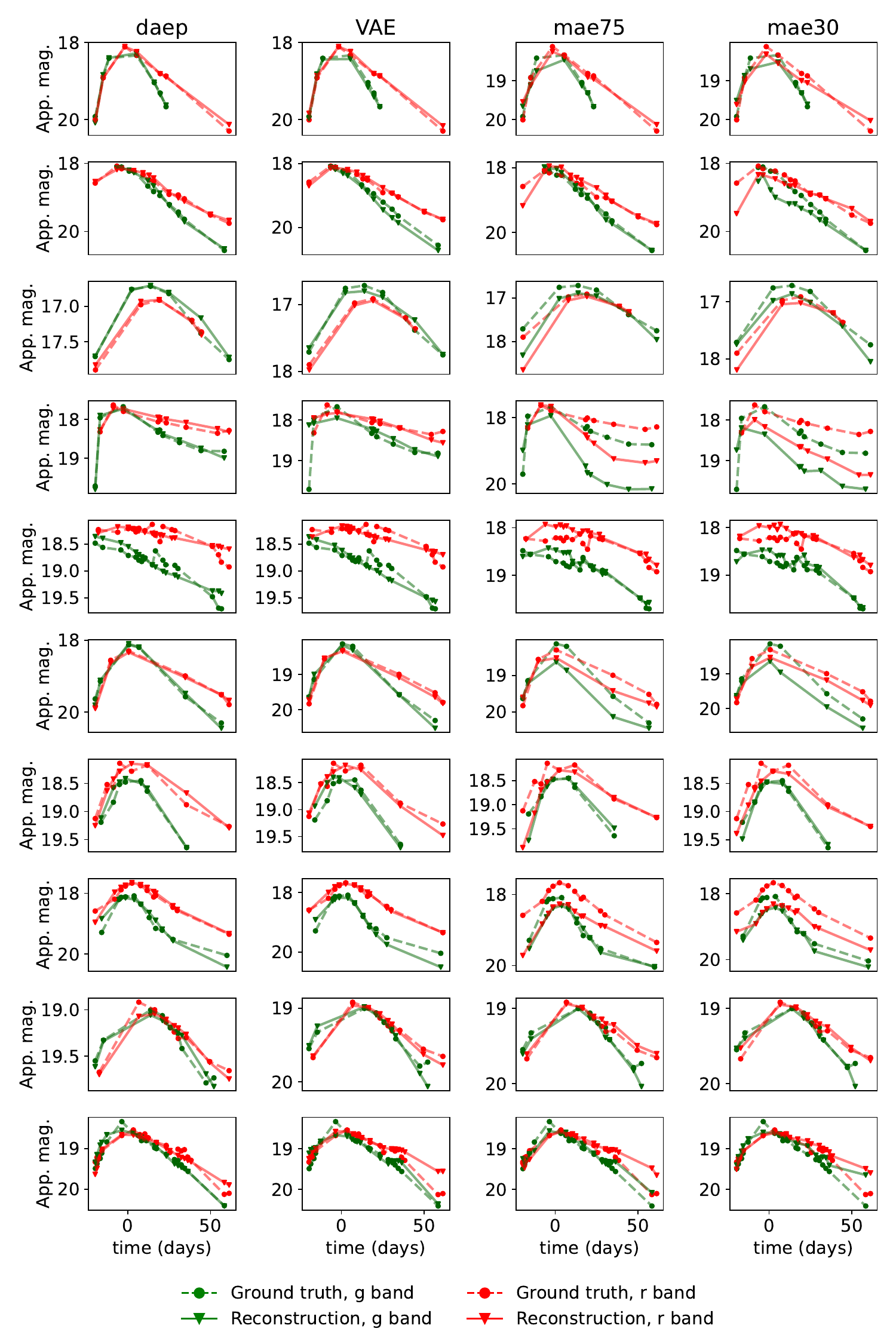}
    \caption{Additional examples of ZTF light curve reconstructions.}
    \label{fig:ztflcvae}
\end{figure}

In \cref{fig:ztflcvae}, we show a series of ZTF light curve reconstructions for daep alongside the VAE and MAE baselines. Our method performs superior reconstructions compared to the baseline models, particularly the MAE with 75\% of the input data masked.  

\begin{figure}[htp]
    \centering
    \includegraphics[width = \linewidth]{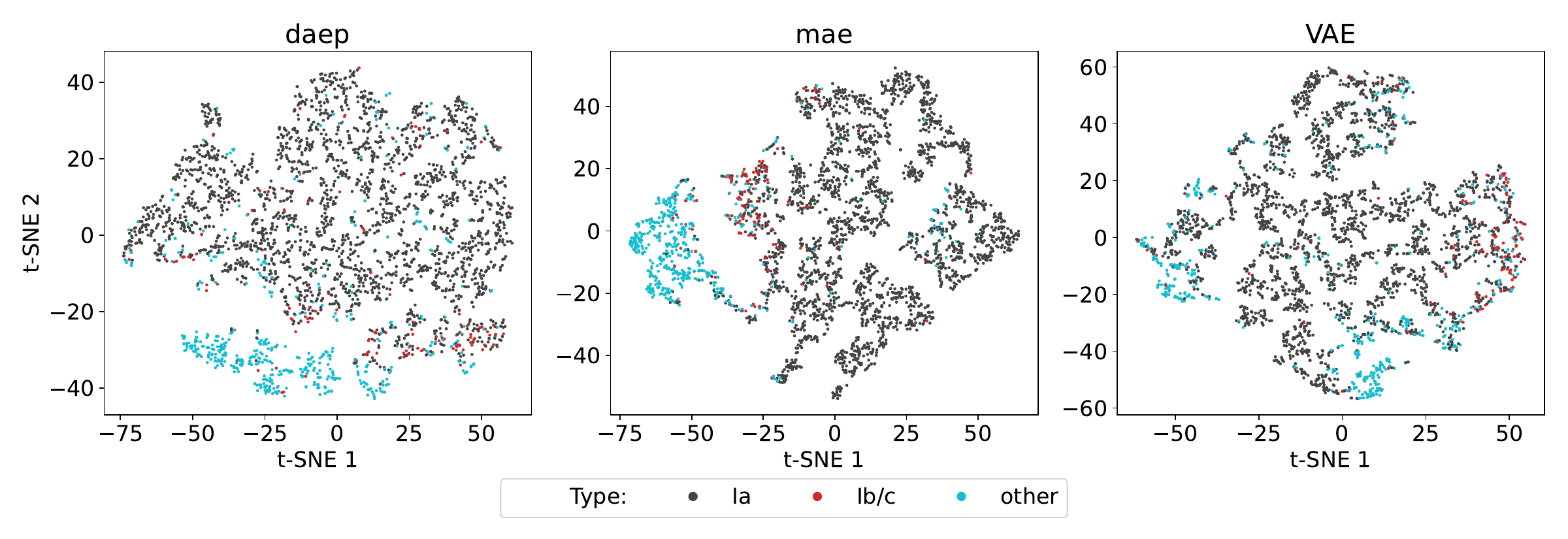}
    \includegraphics[width = \linewidth]{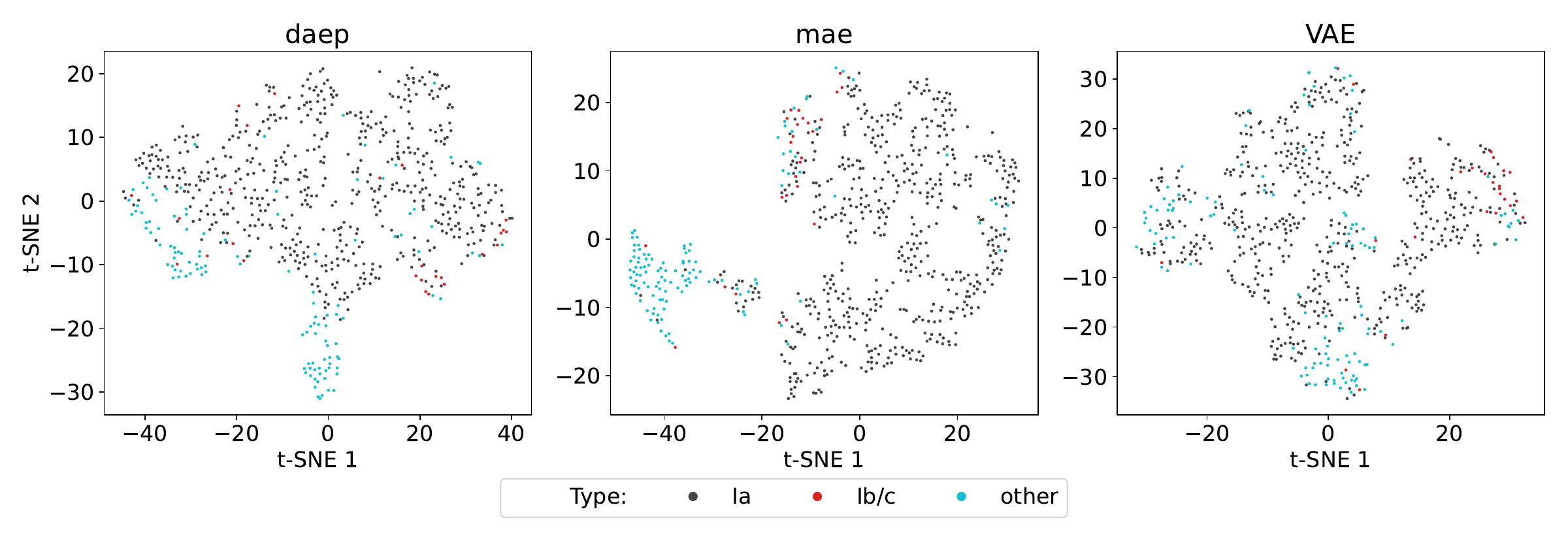}
    \caption{Latent representations of ZTF photometry from training (upper) and test (lower). The latent space for daep appears more well-regularized compared to a VAE of comparable dimensionality.}
    \label{fig:ztflcencode}
\end{figure}

In \cref{fig:ztflcencode}, we show the light curve latent representations after t-SNE for all three models considered in this work. As with the ZTF spectra, daep's latent space appears more continuous than either MAE/VAE baselines (though type~Ib/c supernovae do not appear as well-separated as in the MAE space, as indicated by the higher mean F$_1$ score from mae30 listed in \cref{tab:metricZTFphotometry}.).

\end{document}